\documentclass[sigconf]{acmart}

\usepackage{booktabs} 

\usepackage{url}
\usepackage{graphicx}  
\usepackage{amsfonts} 
\usepackage{CJK}
\usepackage{amsmath}
\usepackage{bm}
\usepackage{multirow}
\usepackage{subfigure}
\usepackage{algorithm}  
\usepackage{algpseudocode}  
\usepackage{makecell}
\usepackage{xspace}
\usepackage{amssymb}
\usepackage{array}
\usepackage[normalem]{ulem}
\usepackage{footmisc}
\usepackage{listings}
\usepackage[np, autolanguage]{numprint}
\usepackage{paralist}
\usepackage{enumitem}
\usepackage{colortbl}
\usepackage{soul}
\usepackage{epstopdf}
\usepackage{setspace}
\usepackage{pifont}
\usepackage{xcolor}
\usepackage{mathrsfs}

\newcommand{\ie}{\emph{i.e.,}\xspace}

\newcommand{\eg}{\emph{e.g.,}\xspace}
\newcommand{\etal}{\emph{et al.}\xspace}
\newcommand{\ignore}[1]{}

\newcommand{\dubbelop}{$^{\blacktriangle}$}

\newcommand{\dubbelneer}{$^{\blacktriangledown}$}

\newcommand{\norm}[1]{\left\lVert#1\right\rVert}
\copyrightyear{2019}
\acmYear{2019}
\setcopyright{acmcopyright}
\acmConference[WSDM '19]{The Twelfth ACM International Conference on Web Search and Data Mining}{February 11--15, 2019}{Melbourne, VIC, Australia} 
\acmBooktitle{The Twelfth ACM International Conference on Web Search and Data Mining (WSDM '19), February 11--15, 2019, Melbourne, VIC, Australia} 
\acmPrice{15.00}
\acmDOI{10.1145/3289600.3290992} 
\acmISBN{978-1-4503-5940-5/19/02}
\title{Product-Aware Answer Generation \\in E-Commerce Question-Answering}
\begin{document}



\author{Shen Gao}
 \authornote{Work performed during an internship at JD.com.}
\affiliation{%
  \institution{ICST, Peking University}
}
\email{shengao@pku.edu.cn}

\author{Zhaochun Ren}
\affiliation{%
  \institution{JD.com}
}
\email{renzhaochun@jd.com}

\author{Yihong Zhao}
\affiliation{%
	\institution{JD.com}
}
\email{ericzhao@jd.com}

\author{Dongyan Zhao}
\affiliation{%
	\institution{ICST, Peking University}
}
\email{zhaody@pku.edu.cn}

\author{Dawei Yin}
\affiliation{%
	\institution{JD.com}
}
\email{yindawei@acm.org}

\author{Rui Yan}
\authornote{This author is the corresponding author}
\affiliation{%
  \institution{ICST, Peking University}
}
\email{ruiyan@pku.edu.cn}

\begin{abstract}
In e-commerce portals, generating answers for product-related questions has become a crucial task. 
In this paper, we propose the task of \emph{product-aware answer generation}, which tends to generate an accurate and complete answer from large-scale unlabeled e-commerce reviews and product attributes.
Unlike existing question-answering problems, answer generation in e-commerce confronts three main challenges: (1) Reviews are informal and noisy; (2) joint modeling of reviews and key-value product attributes is challenging; (3) traditional methods easily generate meaningless answers.
To tackle above challenges, we propose an adversarial learning based model, named PAAG, which is composed of three components: a question-aware review representation module, a key-value memory network encoding attributes, and a recurrent neural network as a sequence generator. 
Specifically, we employ a convolutional discriminator to distinguish whether our generated answer matches the facts.
To extract the salience part of reviews, an attention-based review reader is proposed to capture the most relevant words given the question.
Conducted on a large-scale real-world e-commerce dataset, our extensive experiments verify the effectiveness of each module in our proposed model.  Moreover, our experiments show that our model achieves the state-of-the-art performance in terms of both automatic metrics and human evaluations. 

\end{abstract}

%
%




\maketitle

\section{Introduction}
\label{sec:intro}
In recent years, the explosive popularity of \emph{question-answering} (QA) is revitalizing the task of \emph{reading comprehension} with promising results~\cite{wang2017gated,Song2017Summarizing}.
Unlike traditional knowledge-based QA methods that require a structured knowledge graph as the input and output resource description framework (RDF) triples~\cite{He2017GeneratingNA}, most of reading comprehension approaches read context passages and extract text spans from input text as answers~\cite{Seo2017bi, wang2017gated}.

E-commerce are playing an increasingly important role in our daily life.
As a convenience of users, more and more e-commerce portals provide community question-answering services that allow users to pose product-aware questions to other consumers who purchased the same product before.
Unfortunately, many product-aware questions lack of proposer answers.
Under the circumstances, users have to read the product's reviews to find the answer by themselves. Given product attributes and reviews, an answer is manually generated following a cascade procedure:
(1) a user skims reviews and finds relevant sentences; (2) she/he extracts useful semantic units; (3) and the user jointly combines these semantic units with attributes, and writes a proper answer.
However, the information overload phenomenon makes this procedure an energy-draining process to pursue an answer from a rapidly increasing number of reviews.
Consequently, automatic product-aware question-answering become more and more helpful in this scenario.
The task on which we focus is the \emph{product-aware answer generation} given reviews and product attributes.
Our goal is to respond product-aware questions automatically given a large amount of reviews and attributes of a specific product.
Unlike either a ``yes/no'' binary classification task~\cite{McAuley2016Addressing} or a review ranking task~\cite{Moghaddam2011AQAAO}, product-aware answer generation provides a natural-sounding sentence as an answer.

The definition of our task is similar as the reading comprehension. 
However, most of existing reading comprehension solutions only extract text spans from contextual passages~\cite{wei2018fast}.
Since the target of product-aware answer generation is to generate a natural-sounding answer instead of text spans, most of reading comprehension methods and datasets (\eg SQuAD~\cite{Rajpurkar2016SQuAD10}) are not applicable. 
As far as we know, only few of reading comprehension approaches aim to generate a natural-sounding answers from extraction results~\cite{Tan2018snet,mitra2017abstractive}.
With a promising performance on MS-MARCO~\cite{Nguyen2016MSMA}, \emph{S-Net} framework proposed by~\citet{Tan2018snet} focuses on synthesizing answers from extraction results. However, S-Net requires a large amount of labeling data for extracting text spans, which is still unrealistic given a huge number of reviews.
Moreover, product reviews from e-commerce website are informal and noisy, whereas in reading comprehension the given context passages are usually in a formal style. 
Generally, existing reading comprehension approaches confront three challenges when addressing product-aware question answering:
(1) Review text is irrelevant and noisy.
(2) It's extremely expensive to label large amounts of explicit text spans from real-world e-commerce platforms. 
(3) Traditional loss function calculation in reading comprehension tends to generate meaningless answers such as ``I don't know''.

In this paper, we propose the \textit{product-aware answer generator} (PAAG), a product related question answering model which incorporates customer reviews with product attributes. Specifically, at the beginning we employ an attention mechanism to model interactions between a question and reviews.
Simultaneously, we employ a key-value memory network to store the product attributes and extract the relevance values according to the question.
Eventually, we propose a recurrent neural network (RNN) based decoder, which combines product-aware review representation and attributes to generate the answer.
More importantly, to tackle the problem of meaningless answers, we propose an adversarial learning mechanism in the loss calculation for optimizing parameters.
Conducted on a large-scale real-world e-commerce dataset, we evaluate the performance of PAAG using extensive experiments.
Experimental results demonstrate that the PAAG model achieves significant improvement over other baselines, including the state-of-the-art reading comprehension model.
Furthermore, we also examine the effectiveness of each module in PAAG. Our experiments verify that adversarial learning is capable to significantly improve the denoising and facts extracting capacity of PAAG.

\smallskip\noindent
\noindent To sum up, our contributions can be summarized as follows:

$\bullet$ We propose a product-aware answer generation task.

$\bullet$ To tackle this task, we propose an end-to-end learning method to extract fact that is helpful for answering questions from reviews and attributes and then generate answer text.

$\bullet$ Due to the review is in an informal style with noise, we propose an attention based review reader and use the Wasserstein distance based adversarial learning method to learn to denoise the review text. The discriminator can also give an additional training signal for generating more consistence answer.

$\bullet$ Experiments conducted on a large-scale real-world dataset show that our PAAG method outperforms all baselines, include the state-of-the-art model in terms of all metrics.
The effectiveness of each module in PAAG is also demonstrated in our experiments.

\section{Related Work}

We detail related work on product-aware question-answering, reading comprehension, and sequence-to-sequence architecture.

\noindent \textbf{Product-aware question answering.} In recent years, product-aware question answering has received several attention. Most of existing strategies aim at extracting relevant sentences from input text to answer the given question.
Yu \etal~\cite{yu2012answering} propose a framework for opinion QA, which first organizes reviews into a hierarchy structure and retrieves review sentence as the answer. Yu \etal~\cite{yu2018aware} propose an answer prediction model by incorporating an aspect analytic model to learn latent aspect-specific review representation for predicting the answer.
External knowledge has been considered with the development of knowledge graphs.
McAuley \etal~\cite{McAuley2016Addressing} propose a method using reviews as knowledge to predict the answer, where they classify answers into two types, binary answers (i.e. ``yes'' or ``no'') and open-ended answers.
Incorporating review information, recent studies employ ranking strategies to optimize an answer from candidate answers~\cite{Yu2018ModellingDR,Moghaddam2011AQAAO}. 
Meanwhile, product-aware question retrieval and ranking has also been studied.
Cui \etal~\cite{Cui2017SuperAgentAC} propose a system which combines questions with RDF triples.
Yu \etal~\cite{Yu2018ModellingDR} propose a model which retrieves the most similar queries from candidate QA pairs, and uses corresponding answer as the final result.

\noindent However, all above task settings differ from our task.
Unlike above approaches, our method is aimed to generate an answer from scratch, based on both reviews and product attributes.

\noindent \textbf{Reading Comprehension.} Given a question and relevant passages, reading comprehension extracts a text span from passages as an answer~\cite{Rajpurkar2016SQuAD10}. Recently, based on a widely applied dataset, i.e., SQuAD~\cite{Rajpurkar2016SQuAD10}, many appraoches have been proposed.
Seo \etal~\cite{Seo2017bi} use bi-directional attention flow mechanism to obtain a query-aware passage representation.
Wang \etal~\cite{wang2017gated} propose a model to match the question with passage using gated attention-based recurrent networks to obtain the question-aware passage representation.
Consisting exclusively of convolution and self-attention, QANet~\cite{wei2018fast} achieves the state-of-the-art performance in reading comprehension.
As mentioned above, most of the effective methods contain question-aware passage representation for generating a better answer.
This mechanism make the models focus on the important part of passage according to the question.
Following these previous work, our method models the reviews of product with a question aware mechanism.

\noindent \textbf{Sequence-to-sequence architecture.} In recent years, sequence-to-sequence (seq2seq) based neural networks have been proved effective in generating a fluent sentence.
The seq2seq model~\cite{Sutskever2014SequenceTS} is originally proposed for machine translation and later adapted to various natural language generation tasks, such as text summarization and dialogue generation~\cite{Tao2018RUBERAU,Yao2017TowardsIC}.
Rush \etal~\cite{Rush2015ANA} apply the seq2seq mechanism with attention model to text summarization field.
Then See \etal~\cite{See2017GetTT} add copy mechanism and coverage loss to generate summarization without out-of-vocabulary and redundancy words.
The seq2seq architecture has also been broadly used in dialogue system.
Tao \etal~\cite{Tao2018Get} propose a multi-head attention mechanism to capture multiple semantic aspects of the query and generate a more informative response.
Different from seq2seq models, our model utilizes not only the information in input sequence but also many external knowledge from user reviews and product attributes to generate the answer that matches the facts.
Unlike traditional seq2seq model, there are several tasks which input data is in key-value structure instead of a sequence.
In order to utilize these data when generating text, key-value memory network (KVMN) is purposed to store this type of data.
He \etal~\cite{He2017GeneratingNA} incorporate copying and retrieving knowledge from knowledge base stored in KVMN to generate natural answers within an encoder-decoder framework. 
Tu \etal~\cite{Tu2018Learning} use a KVMN to store the translate history which gives model the opportunity to take advantage of document-level information instead of translate sentences in an isolation way.
We will use the KVMN architecture in our model to store and retrieve the product attributes data.

\section{Problem formulation}
\label{sec:formulation}

Before introducing our answer generation task for product-aware question, we introduce our notation and key concepts.

At the beginning, for a product, we assume there is a question $X^q = \{x^q_1, x^q_2, \dots, x^q_{T_q}\}$, $T_r$ reviews $X^r = \{x^r_1, x^r_2, \dots, x^r_{T_r}\}$ and $T_a$ key-value pairs of attributes $A = \{(a^k_1, a^v_1), (a^k_2, a^v_2), \dots, (a^k_{T_a}, a^v_{T_a})\}$, where $a^k_i$ is the name of $i$-th attribute and $a^v_i$ is the attribute content.
In our task, we assume that each attribute, both key $a^k_i$ and value $a^v_i$ are represented as a single word.
Given a question $X^q$, an answer generator reads the reviews $X^r$ and attributes $A$, then generates an answer $\hat{Y} = \{\hat{y}_1, \hat{y}_2, \dots, \hat{y}_{T_y}\}$.
The goal is to generate an answer $\hat{Y}$ that is not only grammatically correct but also consistent with product attributes and opinions in the reviews.
Essentially, the generator tries to optimize the parameters to maximize the probability $P(Y|X^q, X^r, A) = \prod_{t=1}^{T_y} P(y_t|X^q, X^r, A)$ where $Y = \{y_1, y_2, \dots, y_{T_y}\}$ is the ground truth answer.

\section{PAAG model}

\subsection{Overview}

\begin{figure*}
    \centering
    \includegraphics[scale=0.45]{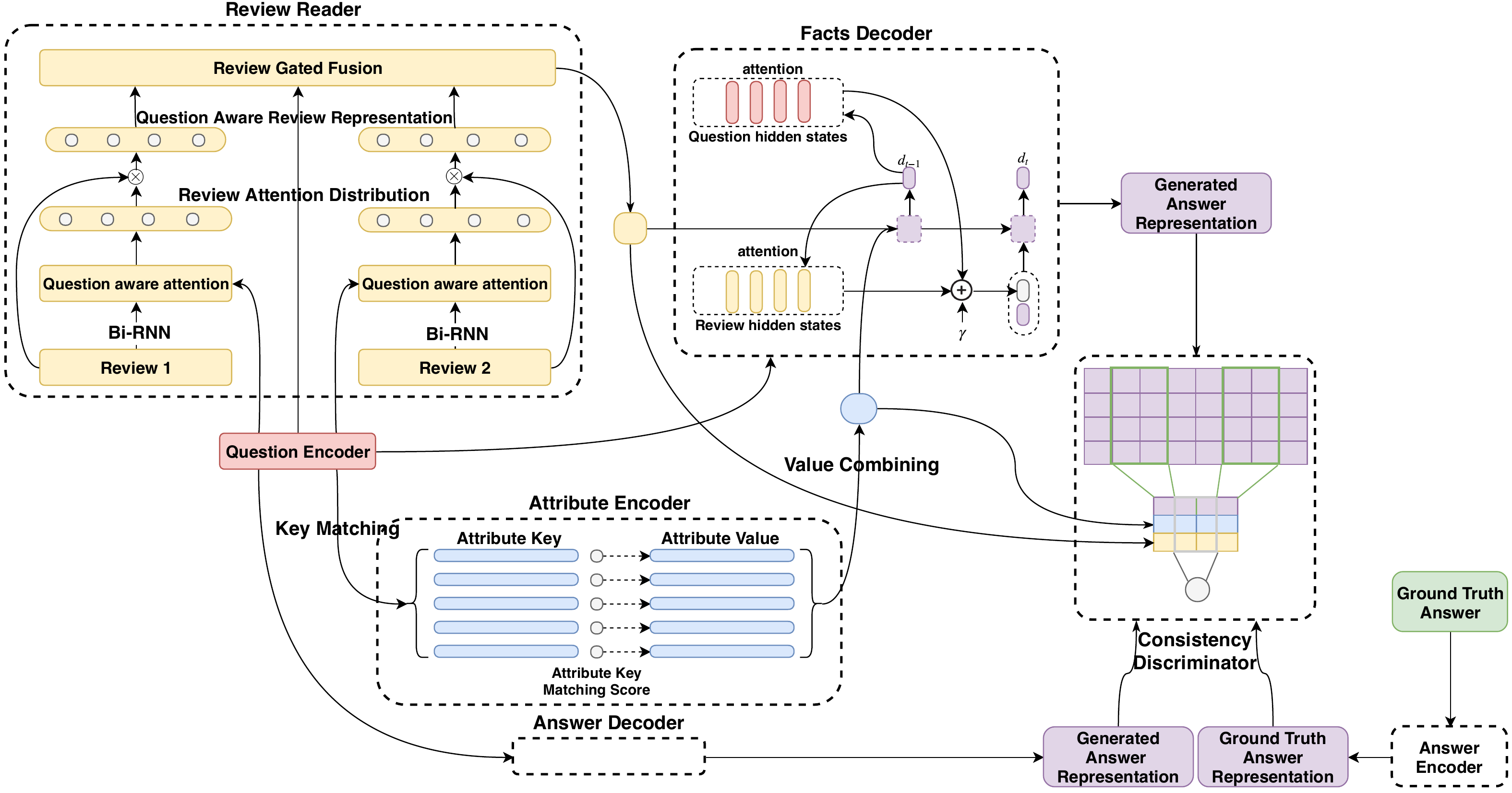}
    \caption{Overview of PAAG. We divide PAAG into four parts: (1) \textit{Review reader} reads the review to extract relevant semantic parts. (2) \textit{Attribute encoder} encodes the attribute key-value pairs using key-value memory network. (3) \textit{Facts decoder} generates the final answer according to the facts learned by the two modules introduced before. (4) \textit{Consistency discriminator} distinguishes whether the generated answer matches the extracted facts, and we also use the result of discriminator as another training signal.}
    \label{fig:overview}
\end{figure*}
In this section, we propose our product-aware answer generator model, abbreviated as PAAG. The overview of PAAG is shown in Figure~\ref{fig:overview}.
PAAG can be split into two main parts: answer generator and consistency discriminator.
We start by detailing the answer generator which generates an answer according to the reviews and attributes.
We then describe the consistency discriminator which distinguishes whether the generated answer matches the facts given by reviews and attributes.

$\bullet$ \textit{Answer generator.}
(1) \textbf{Review reader:} (See Section~\ref{sec:review-reader}) In this part, we encode the review text into vector representations. By matching the relevance of the given question, we signify important semantic units of reviews.
(2) \textbf{Attributes encoder:} (See Section~\ref{sec:attributes-encoder}) Our model stores the product attributes information into a key-value memory network. For each key-value pair, a correlation score between a key and the question is aggregated into the value.
(3) \textbf{Facts decoder:} (See Section~\ref{sec:decoder}) To generate the answer, we use the RNN-based decoder which fuses the facts extracted from reviews and attributes when generating words.

$\bullet$ \textit{Consistency discriminator.} (See Section~\ref{sec:discrimiantor})
Existing approaches easily generate a grammatically correct answer but conflicts to the facts.
In order to produce factual answer, we use a discriminator to determine whether the generated sentence matches the facts.  
By employing the earth-mover distance to optimize our network, we use the result of discriminator as a training signal to encourage our model to produce a better answer.

\subsection{Review reader} \label{sec:review-reader}

At the beginning, we use an embedding matrix $e$ to map one-hot representation of each word in the question $X^q$, reviews $X^r$, and attributes $A$ to a high-dimensional vector space.
We denote $e(x)$ as the embedding representation of word $x$.
From these embedding representations, we employ a bi-directional recurrent neural network (Bi-RNN) to model the temporal interactions between words:
\begin{align}
    h^q_t &= \text{Bi-RNN}_q(e(x^q_t), h^q_{t-1}), \\
    h^r_{i,t} &= \text{Bi-RNN}_r(e(x^r_{i,t}), h^r_{i,t-1}).
\end{align}
\noindent where $h^{q}_t$ and $h^r_{i,t}$ denote the hidden state of $t$-th step in Bi-RNN for question $X^q$ and $i$-th review in $X^r$ respectively. We denote the final hidden state $h^q_{T_q}$ of $\text{Bi-RNN}_q$ as the vector representation of the question $X^q$.
Following~\cite{Wang2018joint, Seo2017bi}, we choose the long short-term memory (LSTM) as a cell of Bi-RNN. 

For producing a fixed size vector representation of reviews, an intuitive method is to conduct an average-pooling strategy on all the hidden states of each review, which neglects the question-oriented salient part of the review.
Accordingly, we propose a gated attention-based method to incorporate the focus point of question into the review representation.
Furthermore, we add an additional gate which learns the relevance between question $X^q$ and review $X^r$ via a soft-alignment, so we have:
\begin{align}
    s^k_{i,j} &= v^\intercal \tanh(W_q h^q_k + W_r h^r_{i,j}), \\ 
    s_{i,j} &= \max({s^1_{i,j}, s^2_{i,j}, \dots, s^{T_q}_{i,j}}), \\ 
    \alpha_{i,j} &= \exp(s_{i,j}) / \textstyle \sum^{T^i_r}_{t=1} \exp(s_{i,t}) ,
\end{align}
\noindent where $W_q, W_r, v$ are all trainable parameters. $\alpha_{i,j} \in \mathbb{R}$ refers to the importance score of the $j$-th word in the $i$-th review given $X^q$.
Thereafter, we apply the attention-pooling operation on each review hidden state $h^r_{i,\cdot}$ to produce the question-aware review representation $c^r_i$, shown in Equation~\ref{rev-attn-pool}:
\begin{align}\label{rev-attn-pool}
    c^r_i = \textstyle \sum^{T^i_r}_{t=1} \alpha_{i,t} h^r_{i,t},
\end{align}

Given an answer generation procedure, not all the reviews are useful to answer the question due to the informal style problem.
Thus if we directly calculate the arithmetic mean vector of all reviews' representations, we can not capture salient passages.
To tackle this problem, a \textbf{gated fusion} method is utilized here to sum up all the review representations.
We first calculate the relevance between each review representation $c^r_i$ and question representation $h^q_{T_q}$ via a bi-linear layer shown in Equation \ref{rev-gate}:
\begin{align}
    u_i &= c^r_i W_f h^q_{T_q} \label{rev-gate},\\
    u^{'}_i &= \exp(u_i) / \textstyle \sum^{T_r}_{t=1} \exp(u_t), \label{rev-gate-softmax}
\end{align}
\noindent where $W_f$ is a trainable parameter. 
Afterwards, we use a softmax function to simulate the relevance score $u_i$, \ie $u^{'}_i$, shown in Equation \ref{rev-gate-softmax}.
Then we use $u^{'}_i$ as the weight of the $i$-th review to do weighted-average on review representation $c^r_i$ over all reviews, so we have:
\begin{equation}
    c^r = \textstyle \sum^{T_r}_{i=1} u^{'}_i c^r_i .
\end{equation}

\subsection{Attributes encoder}\label{sec:attributes-encoder}

The attributes of a product can be seen as structured knowledge data in our task. As key-value memory network (KVMN) is shown effective in structured data utilization~\cite{miller2016key,Tu2018Learning,He2017GeneratingNA}, in our work we employ KVMN to store product attributes for generating answers.
Correspondingly, we store the word embedding of each attribute's key and value in the KVMN.
The \emph{read} operation in our KVMN is divided into two steps: \emph{key matching} and \emph{value combination}.

\textbf{Key matching} 
The goal of key matching is to calculate the relevance between each attribute and the given question. 
Given question $X^q$, for the $i$-th attribute $a_i=(a^k_i, a^v_i) \in A$, we calculate the probability of $a_i$ over $X^q$, \ie $P(a_i | X^q)$, as the \emph{matching score}. To this end, we exploit the question representation $h^q_{T_q}$ to calculate the probability distribution:
\begin{equation} \label{equ:attr-key-score}
    P(a_i | X^q) = \frac{\exp(h^q_{T_q} W_a e(a^k_i))}{\textstyle \sum^{T_a}_{t=1} \exp(h^q_{T_q} W_a e(a^k_t))} ,
\end{equation}

Since question representation $h^q_{T_q}$ and attribute key representation $e(a^k_i)$ are not in the same semantic space, following~\cite{Sukhbaatar2015EndToEndMN, Miller2016KeyValueMN}, we use a trainable key matching parameter $W_a$ to transform these representations into a same space.

\textbf{Value combination} As the relevance between question $X^q$ and attribute $a_i$, the matching score $P(a_i | X^q)$ can help to capture the most relevant attribute for generating a correct answer. Therefore, as shown in Equation~\ref{equ:read-kvmn-value}, the attribute encoder reads the information $m$ from KVMN via summing over the stored attribute values, and guide the follow-up answer generation, so we have: 
\begin{equation}\label{equ:read-kvmn-value}
    m = \textstyle \sum^{T_a}_{i=1} P(a_i | X^q) e(a^v_i) .
\end{equation}

\subsection{Facts decoder} \label{sec:decoder}

PAAG generates an answer based on a set of facts extracted from reviews and attributes. 
Same as our encoder settings, we set LSTM as a cell in our RNN-based facts decoder. We concatenate the question, review and attribute representations and apply a linear transform, then use this vector as the initial state $d_0$; at every decoding step, we feed a context vector $g_t$ into RNN cell.
At $t$-th decoding step, context vector $g_t$ summarizes the input question and review, and we will show the detail of producing $g_t$ at follows.
The procedure of $t$-th decoding step is shown in Equation~\ref{dec-step}.
We use the notion $\left[ \cdot ; \cdot \right]$ as the concatenation of two vectors.
\begin{align}
    d_0 &= W_e \left[m; h^q_{T_q}; c^r\right] + b_e, \\
    d_t &= \text{LSTM} \left( d_{t-1}, \left[ g_{t-1}; e(y_{t-1}) \right] \right) , \label{dec-step}
\end{align}
\noindent where $W_e, b_e$ are the trainable parameters, $d_t$ is the hidden state of $t$-th decoding step.

Similar with the seq2seq with attention mechanism, we use the hidden state of previous step $d_{t-1}$ to attend the question hidden states and review hidden states to get the context vector $g_t$ of current decoding step.
The algorithm of attending reviews hidden states is same as attending question hidden states, so we use $h^{*}_i$ to represent the hidden state where $*$ can be $r$ or $q$.
\begin{align}
    \beta^{'}_{i, t} &= z^\intercal \tanh \left( W_s h^{*}_i + W_d d_t \right), \\ 
    \beta_{i, t} &= \exp \left( \beta^{'}_{i, t} \right) / \textstyle \sum^{T^q}_{j=1} \exp \left(\beta^{'}_{j, t} \right), \\
    g^{*}_t &= \textstyle \sum_{i=1}^{T} \beta_{i, t} h^{*}_i ,
\end{align}
where $W_s, W_d$ are all trainable parameters.
After two attention procedures of question and review finished, we concatenate context vector $g^r_t$ and $g^q_t$ with a balanced gate $\gamma$ which is determined by decoder state $d_t$:
\begin{equation}
    \gamma = \sigma \left( W_g d_t + b_g \right), \enskip g_t = \left[ \gamma g^r_t; \left( 1 - \gamma \right) g^q_t \right],
\end{equation}
The context vector $g_t$, which can be seen as a representation of reading from the question and reviews, is concatenated with the decoder state $d_t$ and then fed into a linear transformation layer to conduct the generated word distribution $P_{v}$ over the vocabulary.
\begin{align}
    d^o_t &= \left(W_o[d_t; g_t]+b_o \right) \label{dec-linear},\\
    P_{v} &= \text{softmax} \left(W_v d^o_t + b_v \right), 
\end{align}

At the $t$-th decoding step, we set the loss as the negative log likelihood of the target word $y_t$:
\begin{equation}\label{loss-g}
    loss_g = - 1/T_y \textstyle \sum^{T_y}_{t=1} \log P_{v}(y_t).
\end{equation}

In order to handle the out-of-vocabulary (OOV) problem, we equip the pointer network~\cite{Gu2016IncorporatingCM,vinyals2015pointer,See2017GetTT} with our decoder, which makes our decoder capable to copy words from question.
The procedure of pointer network is the same as the model proposed by See \etal~\cite{See2017GetTT}, and we omit this procedure in our paper due to the limited space.

Up to now, we can use $loss_g$ to compute gradients for all the parameters in answer generator and use gradient descent method to update these parameters.
But the correctness constraint given by cross entropy loss $loss_g$ is not enough.
So we need a classifier to judge whether the generated answer is consistent with the facts.
In this way, we use this classification result to guide the answer generator to produce more consistent answers.

\subsection{Consistency discriminator}\label{sec:discrimiantor}

To generate sentences which are more consistent with the facts, we add a discriminator to provide additional training signals for the answer generator.
We propose a convolutional neural network (CNN) based classifier as discriminator.
The goal of this classifier is to distinguish whether a sentence is consistent with the given facts.
So we can use the confidence of classifying a sentence as a training signal to encourage the answer generator to produce a better answer.
We use the answer generated by the facts decoder presented in Section~\ref{sec:decoder} as the negative sample for classification, and use the representation of ground truth answer as the positive sample.

As for giving a positive sample for discriminator, we use an RNN to encode the ground truth answer into a vector representation.
\begin{equation}\label{equ:ground-truth-enc}
    \tilde{d}^g_t = LSTM(y_t, \tilde{d}^g_t)
\end{equation}
Since the ground truth is encoded by another RNN which is different from the decoder RNN of $d^o_t$, we use a linear transformation to transform the high-dimensional space of $\tilde{d}^g_t$ to the same space as $d^o_t$ in.
\begin{equation}\label{equ:ground-truth-trans}
    d^g_t = W_z * \tilde{d}^g_t + b_z
\end{equation}
where $W_z$ and $b_z$ are all trainable parameters.

For training the discriminator ability of capturing whether an answer is consistent with the facts, we construct another negative samples to train the discriminator. 
We present an answer decoder (shown at the bottom of Figure~\ref{fig:overview}), which employs the same decoding mechanism as the facts decoder but no fact is attached during decoding.
Specifically, we use the hidden state $d^o_t$ shown in Equation~\ref{dec-linear} as the representation of each word in the generated sentence by the facts decoder.
Similarly, the decoder RNN without feeding facts representation $g_t$ generates hidden states $d^f_t$.
We use $d^f_t$ as the representation of generated answer without facts support.

Then a two-dimensional convolutional layer convolves the hidden states $d^*_t$ with multiple convolutional kernels of different widths. 
Each kernel corresponds a linguistic feature detector which extracts a specific  pattern of multi-grained n-grams~\cite{Kalchbrenner2014ACN}.
A convolutional filter $W_c$ maps hidden states in the receptive field to a single feature. 
As we slide the filter across the whole sentence, we obtain a sequence of new features $n = [n_1, n_2, \dots, n_l]$, shown in Equation~\ref{dis-cnn}:
\begin{equation}\label{dis-cnn}
    n^{*}_t = \text{relu}\left(d^{*}_t \otimes W_c + b_c\right) ,
\end{equation}
\noindent where $W_c, b_c$ are all trainable parameters and $\otimes$ denotes the convolution operation.
For each convolutional filter, the max-pooling layer takes the maximal value among the generated convolutional features $n^f$, $n^o$ and $n^g$ respectively, resulting in a fixed-size vector $N^{f}$, $N^{o}$ and $N^{g}$.
Then we obtain the classification result $D(d^{*}_t) \in \mathbb{R}$ through an interaction between $N^{*}$ and the facts, i.e., attribute representation $m$ and review representation $c^r$. So we have:
\begin{align}\label{dis-mlp}
    D(d^{*}_t) =  W_h \text{relu} \left( N^{*} + m + c^r \right) + b_h ,
\end{align}
\noindent where $W_h, b_h$ are all trainable parameters. Here we apply the
Vanilla generative adversarial network (GAN) with a sigmoid function on the $D(d^{*}_t)$ to produce the classification probability and tries to minimize the Jensen-Shannon divergence between real and generated data distribution.

However, as vanilla GAN often leads to gradient vanishing as the discriminator saturates~\cite{Gulrajani2017ImprovedTO}, which makes the discriminator can not give the correct training signal.
Inspired by previous work~\cite{Arjovsky2017WassersteinG, Gulrajani2017ImprovedTO}, we tackle this problem by minimize the earth-mover (also called Wasserstein-1) distance $W(\mathbb{P}_r, \mathbb{P}_g)$ instead of Jensen-Shannon divergence. 
Informally, given a distribution $\mathbb{P}_r$ of ground truth answer and a distribution $\mathbb{P}_g$ of facts-based answer or answer without facts. 
Then we minimize the cost of transporting mass from $\mathbb{P}_r$ to $\mathbb{P}_g$.
The discriminator $D \in \mathcal{D}$ is a \emph{1-Lipschitz} function, where $\mathcal{D}$ is the set of \emph{1-Lipschitz} functions~\cite{Gulrajani2017ImprovedTO}.

In order to meet the \emph{Lipschitz} constraint of discriminator $D$, we use an alternative way to enforce the \emph{Lipschitz} constraint.
We add a gradient norm of the output of $D$ with respect to its input, which is simply sampled uniformly along a straight line between points sampled from the ground truth representation $d^g_t$ and the facts based output $d^o_t$.
Then our objective function is shown in Equation~\ref{equ:loss-d}.

\begin{align}
    d^{'}_t &= \epsilon d^o_t + (1 - \epsilon) d^g_t , \\
    loss_d = \frac{1}{T_y} \sum^{T_y}_t D(d^{f}_t) &+ D(d^{o}_t) - D(d^{g}_t) + \lambda \left( \norm{\nabla_{d^{'}_t} D(d^{'}_t)}_2 - 1 \right)^2 \label{equ:loss-d} .
\end{align}

\noindent where $\epsilon \sim U[0, 1]$ is a random number and $\lambda$ is a coefficient of gradient penalty term.
Then we can use the optimization methods to update the parameters of discriminator use the loss function $loss_d$.
Meanwhile, we add the $-D(d^{o}_t)$ to the previous defined $loss_g$ in Equation~\ref{loss-g} to encourage the answer generator produce better result.

\section{Experimental setup}

\subsection{Research questions}\label{sec:research-question}

We list four research questions that guide the remainder of the paper: 
\textbf{RQ1}: What is the overall performance of PAAG? Does it outperform state-of-the-art baselines?
\textbf{RQ2}: What is the effect of each module in PAAG? Does the discriminator give a useful training signal to the answer generation module?
\textbf{RQ3}: Is PAAG capable to extract useful information from noisy reviews?
\textbf{RQ4}: What is the performance of PAAG at different data domain? 

\subsection{Dataset}
We collect a large-scale dataset from a real-world e-commerce website, including question-answering pairs, reviews, and product attributes.
This dataset is available at \url{https://github.com/gsh199449/productqa}.
On this website, users can post a question about the product.
Most questions are asking for an experience of user who has already bought the product.
In the collected data, each QA pair is associated with the reviews and attributes of the corresponding product.
We remove all QA pairs without any relevant review and split the whole dataset into training and testing set.
In total, our dataset contains cover 469,953 products and 38 product categories.
The average length of question is 9.03 words and ground truth answer is 10.3 words.
The average number of attribute is 9.0 key-value pairs.
There are 78.74\% of training samples have more than 10 relevant reviews and 75.33\% of training samples have more than 5 attributes.

\subsection{Evaluation metrics}

To evaluate our proposed method, we employ BLEU~\cite{Papineni2002BleuAM} to measure the quality of generated sentence by computing overlapping lexical units (\eg unigram, bigram) with the reference sentence.
We also consider three embedding-based metrics~\cite{forgues2014bootstrapping} (including Embedding Average, Embedding Greedy and Embedding Extreme) to evaluate our model, following several recent studies on text generation~\cite{Serban2017AHL, Xu2017NeuralRG, Tao2018Get}. 
These three metrics compute the semantic similarity between the generated and reference answer according to the word embedding.

Since automatic evaluation metrics may not always consistent with human perception~\cite{Stent2005EvaluatingEM}, we use human evaluation in our experiment. 
Three annotators are invited to judge the quality of 100 randomly sampled answer generated by different models. 
These annotators are all well-educated Ph.D. students and they are all native speakers. 
Two of them have the background of NLP/summarization and another annotator does not major in computer science.
We show human annotators a question, several reviews and attributes of the product along with answers generated from each model.

Statistical significance of observed differences between the performance of two runs are tested using a two-tailed paired t-test and is denoted using \dubbelop (or \dubbelneer) for strong significance for $\alpha = 0.01$. 

\subsection{Comparisons} \label{sec:baselines}

In order to prove the effectiveness of each module in PAAG, we conduct some ablation models shown in Table~\ref{tab:ablations}.

To evaluate the performance of our dataset and the proposed framework, we compare our model with the following baselines:
(1) \textbf{S2SA}: Sequence-to-sequence framework~\cite{Sutskever2014SequenceTS} has been proposed for language generation task. 
We use seq2seq framework which is equipped with attention mechanism~\cite{Bahdanau2015Neural} and copy mechanism~\cite{Gu2016IncorporatingCM} as baseline method. 
The input sequence is question and ground truth output sequence is the answer.
(2) \textbf{S2SAR}: We implement a simple method which can incorporate the review information when generating the answer.
Different from the S2SA, we use an RNN to read all the reviews and concatenate the final state of this RNN with encoder final state as the initial state of decoder RNN.
(3) \textbf{SNet}: S-Net~\cite{Tan2018snet} is a two-stage state-of-the-art model which extracts some text spans from multiple documents context and synthesis the answer from those spans.
Due to the difference between our dataset and MS-MARCO~\cite{Nguyen2016MSMA}, our dataset does not have text span label ground truth for training the evidence extraction module.
So we use the predicted extraction probability to do weighted sum the original review word embeddings, and use this representation as extracted evidence to feed into the answer generation module.
(4) \textbf{QS}: We implement the query-based summarization model proposed by Hasselqvist \etal~\cite{Hasselqvist2017QueryBasedAS}. 
Accordingly, we use product reviews as original passage and answer as a summary.
(5) \textbf{BM25}: BM25 is a bag-of-words retrieval function that ranks a set of reviews based on the question terms appearing in each review.
We use the top review of ranking list as the answer.
(6) \textbf{TF-IDF}: Term Frequency-Inverse Document Frequency is a numerical statistic that is intended to reflect how important a question word is to a review.
We use this statistic to model the relevance between review and question and select the most similar review as the answer of question.

\begin{table}[t]
\centering
\caption{Ablation models for comparison.}
\label{tab:ablations}
\begin{tabular}{@{}l@{~}l}
\toprule
Acronym & Gloss \\
\midrule

RAGF &  \multicolumn{1}{p{6.5cm}}{\small \raggedright \textbf{R}eview reader + \textbf{A}ttributes encoder + \textbf{G}ated \textbf{F}usion}\\

RAGFD &  \multicolumn{1}{p{6.5cm}}{\small RAGF + consistency \textbf{D}iscriminator}\\

RAGFWD &  \multicolumn{1}{p{6.5cm}}{\small RAGF + \textbf{W}asserstein consistency \textbf{D}iscriminator}\\

PAAG &  \multicolumn{1}{p{6.5cm}}{\small RAGFWD + Gradient Penalty }\\
\bottomrule
\end{tabular}
\end{table}

\subsection{Implementation details}

Without using pre-trained embeddings, we randomly initialize the network parameters at the beginning of our experiments.
All the RNN networks have 512 hidden units and the dimension of word embedding is 256.
To produce better answers, we use beam search with beam size 4.
Adagrad~\cite{Duchi2010AdaptiveSM} with learning rate 0.1 is used to optimize the parameters and batch size is 64.
We implement our model using TensorFlow~\cite{abadi2016tensorflow} framework and train our model and all baseline models on NVIDIA Tesla P40 GPU.

\section{Experimental Result}

\subsection{Overall performance}

For research question \textbf{RQ1}, to demonstrate the effectiveness of PAAG, we examine the overall performance in term of BLEU, embedding metrics and human evaluation. 
Table~\ref{tab:comp_bleu_baselines} and Table~\ref{tab:comp_emb_baselines} list performances of all comparisons in terms of two automatic evaluation metrics.
Significant differences are with respect to SNet (row with shaded background).
In these experimental results, we see that PAAG achieves a 111\%, 8\% and 62.73\% increment over the state-of-the-art baseline SNet in terms of BLEU, embedding greedy and consistency score, respectively.
In Table~\ref{tab:comp_emb_baselines}, we see that our PAAG outperforms all the baseline significantly in semantic distance with respect to the ground truth.

For human evaluation, we ask annotators to rate each generated answer according to two aspects: consistency and fluency.
The rating score ranges from 1 to 3, and 3 is the best.
We finally take the average across answers and annotators, as shown in Table~\ref{tab:comp_human_baslines}.
In Table~\ref{tab:comp_human_baslines}, we can see that PAAG outperforms other baseline models in both sentence fluency and consistency with the facts.
We calculate the variance score in Table~\ref{tab:comp_human_baslines}, which shows that annotators agree with each other's judgments in most cases.
Although the BLEU score of S2SAR is lower than the S2SA, the embedding score and human score for S2SAR are higher than S2SA.
Regardless of few word overlapping between generated and ground answer, the human evaluation and results in terms of embedding metrics verify S2SAR outperforms S2SA.
This observation demonstrates the effectiveness of incorporating review in answer generation.

To explore the difficulty of this task, we use a very intuitive method by adding the review information into decoder shown in S2SAR.
Although there is a small increment of S2SAR with respect to S2SA in all metrics, we still find a noticeable gap between S2SAR and PAAG.
This observation demonstrates that PAAG makes better use of review and attribute information than the simple method S2SAR.
In view of the facts extracted from the review and attributes, we examine directly using the most similar review to question as the answer.
More specifically, we evaluate the performance of the top of review ranking list which is ranked by text similarity algorithm such as BM25 and TF-IDF.
From the result of three metrics, the performance of extractive methods is worth than all the generative methods.
It is worth noting that since the answer generated by extractive methods is written by human, it have very high fluency scores.
But these answers may not match the question, so the consistency score is very low.
Consequently, using the most similar review to question as answer is not a better method than generating answers from scratch.

As our task definition and query based text summarization have some similarities in some way, we can see the reviews as original passage and answer as a query based summary.
We also use the query-based text summarization algorithm~\cite{Hasselqvist2017QueryBasedAS} to generate answer.
Similarly, we also employ a reading comprehension method SNet to tackle this task.
Since query-based text summarization and reading comprehension models are not defined to tackle QA task in e-commerce scenario, it can not fully utilize the interactions between question, review, and attributes.
These methods also lack of ability of denoising the reviews.

\newcommand{\cbkgrnd}{\cellcolor{blue!15}}
\newcommand{\phantomtriangle}{\phantom{\dubbelop}}
\begin{table}[t]
\centering
\caption{BLEU scores comparison between baselines.}
\footnotesize
\begin{tabular}{@{}lcc cc c@{}}
\toprule
& BLEU & BLEU1  & BLEU2 & BLEU3 & BLEU4 \\
\midrule
\multicolumn{6}{@{}l}{\emph{Text generation methods}}\\

S2SA & 1.6186\phantom{0} & 15.4754\phantom{0}& 3.1437\phantom{0} & 0.8267\phantom{0}   & 0.1706\phantom{0} \\

S2SAR & 1.7549 & 15.1708 & 3.2156 & 0.9078 & 0.2142 \\
\cbkgrnd SNet & \cbkgrnd 0.9550 & \cbkgrnd 13.7029 & \cbkgrnd 2.5374 & \cbkgrnd 0.4007 & \cbkgrnd 0.0597  \\
QS & 1.6848 & 15.4961 & 2.9508 & 0.8315 & 0.2119 \\

PAAG & \textbf{2.0189}\dubbelop & \textbf{16.2232}\dubbelop  & \textbf{3.5711}\dubbelop & \textbf{1.0290}\dubbelop & \textbf{0.2787}\dubbelop \\
\midrule
\multicolumn{6}{@{}l}{\emph{Sentence extraction methods}}\\
BM25 & 0.4125 & 6.9630 & 0.7097 & 0.1333 & 0.0439 \\
TF-IDF & 0.2548 & 5.5480 & 0.5127 & 0.0779 & 0.0190 \\
\bottomrule
\end{tabular}
 \vspace{-3mm}
\label{tab:comp_bleu_baselines}
\end{table}

\begin{table}[t]
\centering
\footnotesize
\caption{Embedding scores comparison between baselines.}
\begin{tabular}{@{}lcc c@{}}
\toprule
& Average & Greedy  & Extrema \\
\midrule
\multicolumn{4}{@{}l}{\emph{Text generation methods}}\\

S2SA & 0.410013\phantom{0} & 98.653415\phantom{0} & 0.269461\phantom{0} \\

S2SAR & 0.419979\phantom{0} & 99.742679\phantom{0}& 0.278666\phantom{0} \\
\cbkgrnd SNet & \cbkgrnd 0.397162 & \cbkgrnd 95.791356 & \cbkgrnd 0.277781 \\
QS & 0.400291 & 93.255031 & 0.252164 \\

PAAG & \textbf{0.424218}\dubbelop & \textbf{103.912364}\dubbelop & \textbf{0.288321}\dubbelop \\
\midrule
\multicolumn{4}{@{}l}{\emph{Sentence extraction methods}}\\
BM25 & 0.325946 & 76.814465 & 0.172976 \\
TF-IDF & 0.308293 & 85.020442 & 0.155390 \\
\bottomrule
\end{tabular}
 \vspace{-3mm}
\label{tab:comp_emb_baselines}
\end{table}

\begin{table}[t]
\centering
\caption{Consistency and fluency comparison by human evaluation.}
\footnotesize
\begin{tabular}{@{}lcc cc@{}}
\toprule
& \multicolumn{2}{c}{Fluency} & \multicolumn{2}{c}{Consistency} \\ \cline{2-5} 
& mean & variance  & mean & variance \\
\midrule
\multicolumn{5}{@{}l}{\emph{Text generation methods}}\\
S2SA & 2.22 & 0.3 & 1.62 & 0.29 \\
S2SAR & 2.405 & 0.365 & 1.82 & 0.39 \\
\cbkgrnd SNet & \cbkgrnd 1.93 & \cbkgrnd 0.36 & \cbkgrnd 1.355 & \cbkgrnd 0.225 \\
QS & 2.335 & 0.285 & 1.725 & 0.355 \\
PAAG & 2.865\dubbelop & 0.105 & 2.205\dubbelop & 0.445 \\
\midrule
\multicolumn{5}{@{}l}{\emph{Sentence extraction methods}}\\
BM25 & 2.70 & 0.24 & 1.45 & 0.29 \\
TF-IDF & 2.48 & 0.38 & 1.14 & 0.12 \\
\bottomrule
\end{tabular}
 \vspace{-3mm}
\label{tab:comp_human_baslines}
\end{table}

\begin{table}[t]
\centering
\caption{BLEU scores of different ablation models.}
\footnotesize
\begin{tabular}{@{}lcc cc c@{}}
\toprule
& BLEU & BLEU1  & BLEU2 & BLEU3 & BLEU4 \\
\midrule

RAGF & 1.7931 & 15.7213 & 3.3705 & 0.9385 & 0.2079 \\
RAGFD & 1.8597 & 15.9021 & 3.4160 & 0.9409 & 0.2340 \\
RAGFWD & 1.9389 & 16.1755 & \textbf{3.5986} & 0.9865 & 0.2461 \\
PAAG & \textbf{2.0189} & \textbf{16.2232}  & 3.5711 & \textbf{1.0290} & \textbf{0.2787} \\
\bottomrule
\end{tabular}
 \vspace{-3mm}
\label{tab:comp_bleu_ablation}
\end{table}

\begin{figure}[!t]
  \centering
  \subfigure{
    \label{fig:qa-attn-1}
    \includegraphics[width=8.5cm, height=3.3cm]{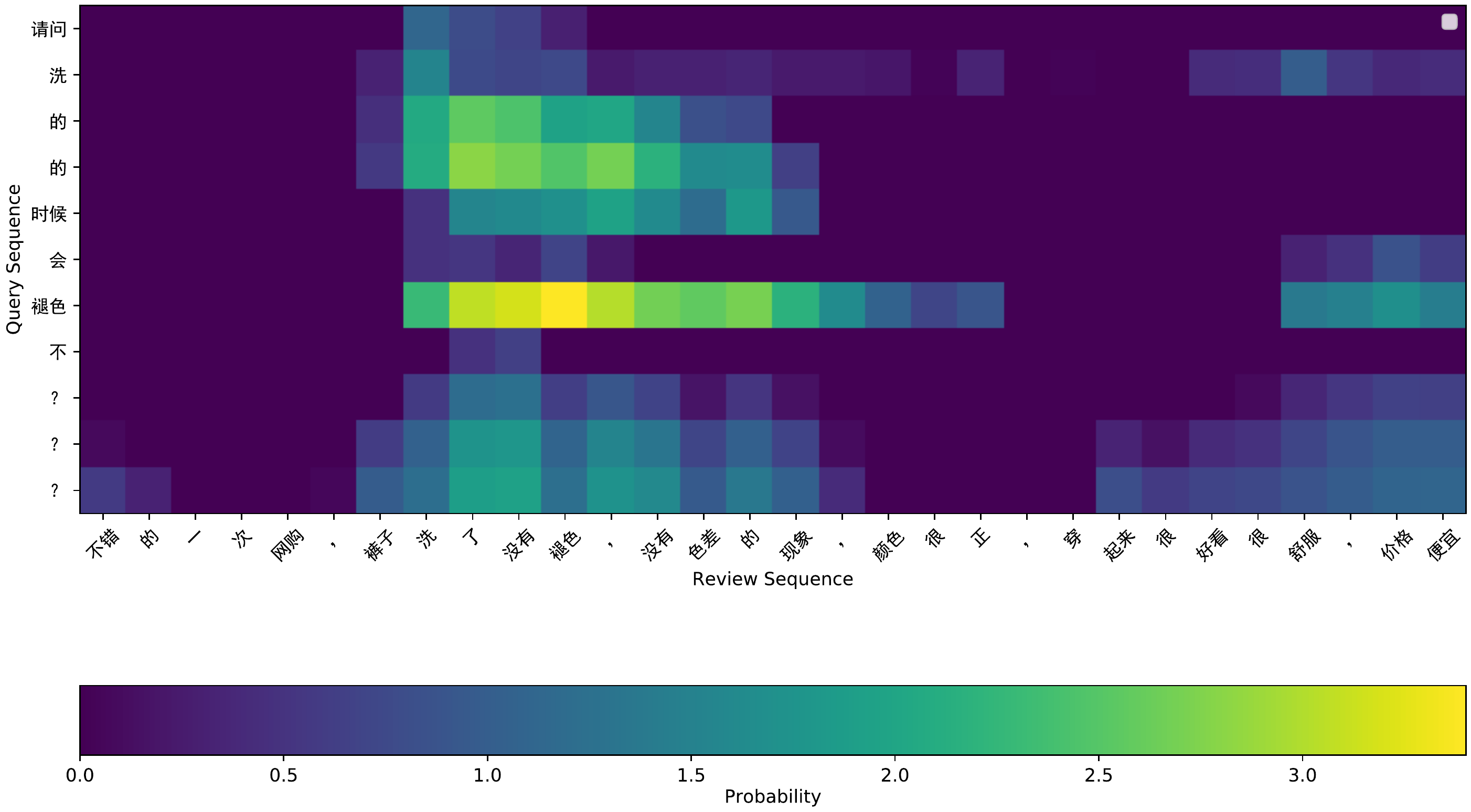}}
  \subfigure{
    \label{fig:qa-attn-2}
    \includegraphics[width=8.5cm, height=2.0cm]{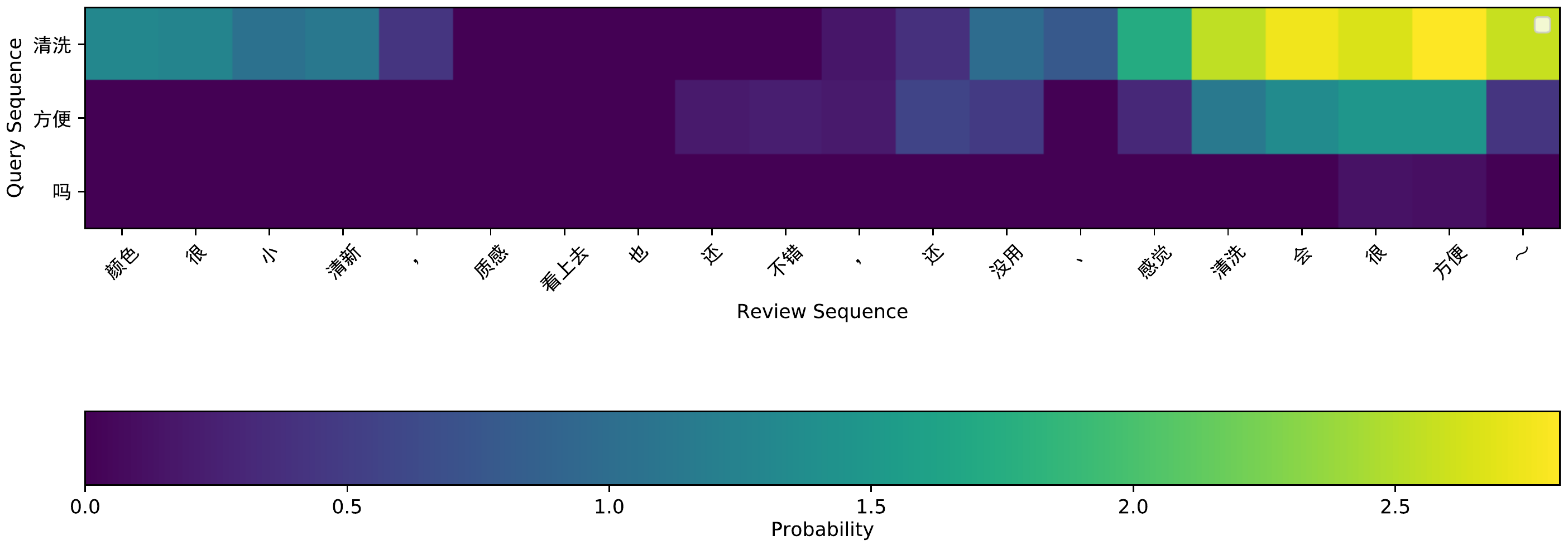}}
    \vspace{-8mm}
    \caption{Visualizations of question-aware review attention map.}
    \vspace{-2mm}
  \label{fig:attention}
\end{figure}

\subsection{Ablation studies}

Next we turn to research question \textbf{RQ2}.
We conduct ablation tests on the usage of adversarial learning method.
The BLEU score of each ablation model is shown in Table~\ref{tab:comp_bleu_ablation}.
In the method RAGFD, we use the vanilla GAN architecture which minimize the divergence.
There is a slight increment from RAGF to RAGFD, which demonstrates the effectiveness of discriminator.
From Table~\ref{tab:comp_bleu_ablation}, we find that RAGFWD achieves a 4.3\% improvement over RAGFD in terms of BLEU, and PAAG outperforms RAGFWD 4.1\% in terms of BLEU.
Accordingly, we conclude that the performance of PAAG benefits from using Wasserstein distance based adversarial learning with gradient penalty.
This approach can help our model to achieve a better performance than the model using the vanilla GAN architecture.

\subsection{Denoising ability}

To address \textbf{RQ3}, in this section we provide an analysis of the denoising ability of our model. According to Table~\ref{tab:comp_bleu_baselines} and Table~\ref{tab:comp_bleu_ablation}, we observe RAGF achieves 2.1\% improvement over SASAR, in terms of BLEU.
Such observation demonstrates that question-aware review generation module gives the denoising ability to the model.
To further investigate the effectiveness of extracting facts from reviews, we visualize two question-review attention maps, shown in Figure~\ref{fig:attention}.
Question of the left figure in Figure~\ref{fig:attention} is ``Will the color fade when cleaning?'' and the right is ``Is it convenient to clean''. 
The review of the left figure is ``Good shopping experience. The pants were washed without discoloration and no color difference compared to the picture. It looks good, comfortable and cheap.'' and the right is ``The color looks good and the texture is great. I haven't started it yet, but it's very easy to clean''.
In this figure, we can see that there is a very strong interaction between question word \begin{CJK*}{UTF8}{gbsn}清洗\end{CJK*} (cleaning) and phrase in review \begin{CJK*}{UTF8}{gbsn}清洗会很方便\end{CJK*} (very easy to clean).
Concretely, these figures show that the question-review attention module can capture the salience semantic part in review according to the question.

\begin{figure}[htbp]
\begin{minipage}[t]{0.49\linewidth}
    \includegraphics[width=\linewidth]{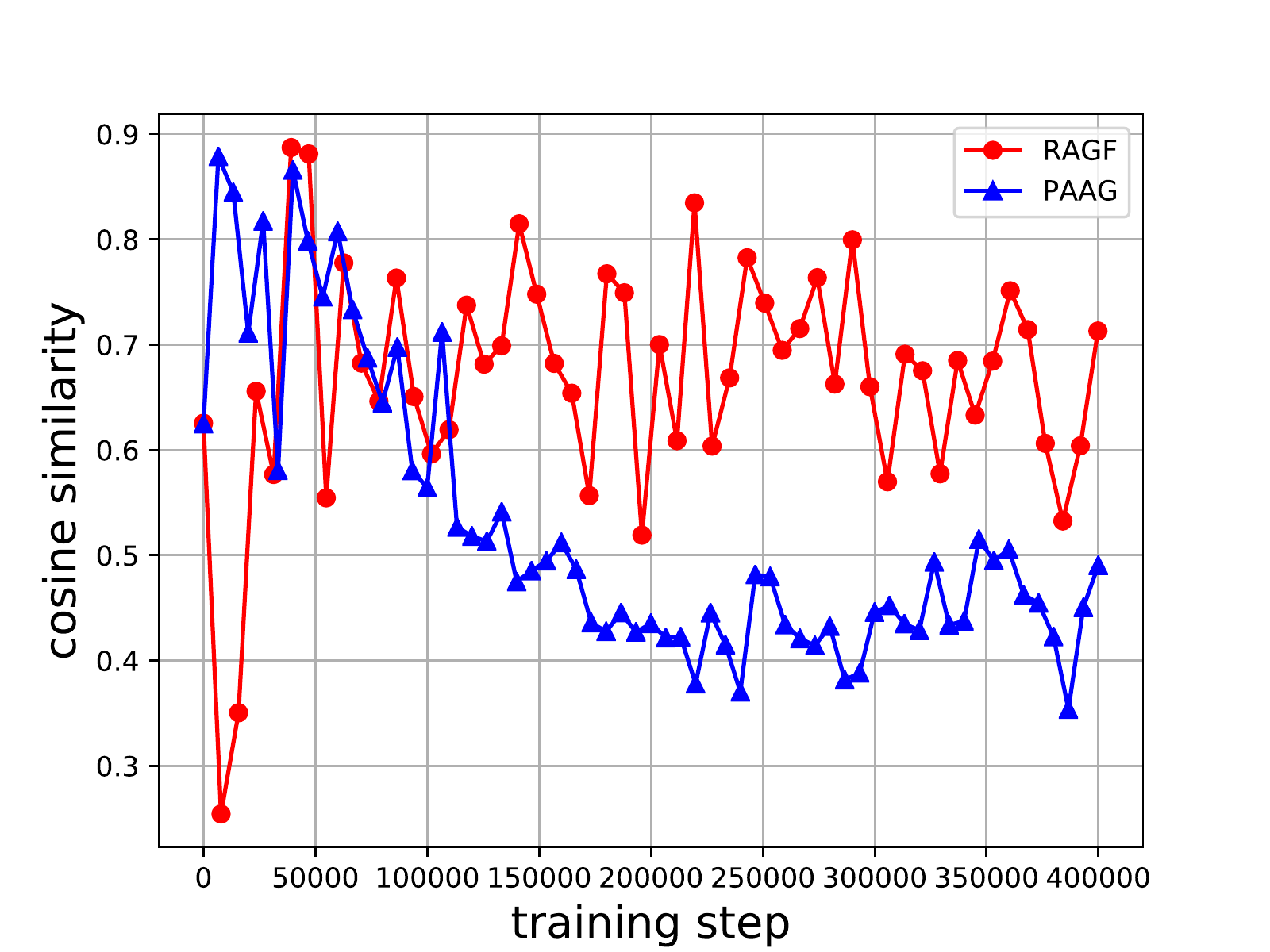}
    \vspace{-7mm}
    \caption{Similarity between review gates and BM25 score.}
    \label{fig:review_gates_simi}
\end{minipage}%
\vspace{-2mm}
    \hfill%
\begin{minipage}[t]{0.49\linewidth}
    \includegraphics[width=\linewidth]{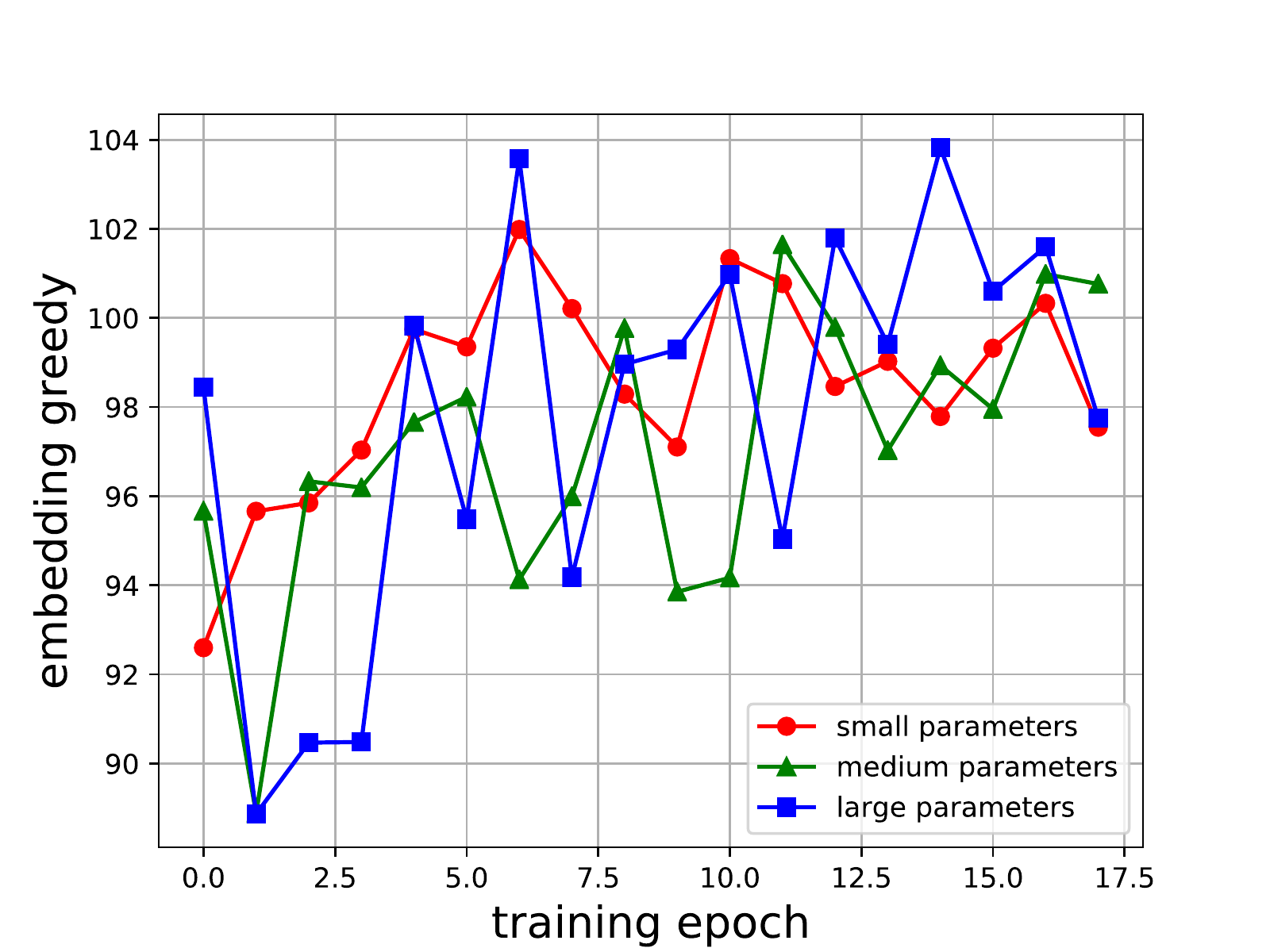}
    \vspace{-7mm}
    \caption{Greedy embedding metric with training epoch.}
    \label{fig:parameter_test}
\end{minipage} 
\vspace{-2mm}
\end{figure}

In the most cases, the higher word overlap between question and review, the more useful the review is.
To prove the ability of review gated fusion module shown in Equation~\ref{rev-gate-softmax}, we use the BM25 algorithm to calculate the similarity between question and each review.
Then we calculate the cosine distance between the salience score produced by review gated fusion module calculated and BM25 similarity score, shown in Figure~\ref{fig:review_gates_simi}.
In order to demonstrate the denoising ability of adversarial learning method, we compare our full model PAAG with the baseline model RAGF, this experiment proves that the usage of WGAN can encourage our model to capture the salience review better.

\vspace{-2mm}

\subsection{Discussions}

\begin{CJK*}{UTF8}{gbsn}
\begin{table}[t]
\centering
\caption{Examples of the generated natural answers by PAAG and other models.}
\scriptsize
\vspace{-5mm}
\begin{tabular}{l|l}
\toprule
\multicolumn{1}{c|}{\multirow{6}{*}{reviews}} & \multicolumn{1}{p{6.5cm}}{衣服 质量 很 好 ， 就 是 我 比较 瘦 ， 这个 S 号 对于 我 来说 还是 比较 肥 。 很 适合 孕妇 穿 。 (The quality of the clothes is very good. Because I am thin, the S size is still quite fat for me. It is suitable for pregnant women to wear.)}     \\ \cline{2-2} 
\multicolumn{1}{c|}{}                         & \multicolumn{1}{p{6.5cm}}{颜色 漂亮 ， 宽松 舒服 ， 就 是 线头 有点 多 ， 竟然 还有 口袋 ， 方便 实用 ！ (The clothes are beautiful in color, comfortable to wear, and the thread is a bit more. This dress has a pocket and it is convenient and practical!)}      \\ \cline{2-2} 
\multicolumn{1}{c|}{}                         & \multicolumn{1}{p{6.5cm}}{不 轻 的 雪纺 连衣裙 ， 很 有 垂坠感 ， 觉得 夏天 春天 都 适合 穿 ， 很 衬肤 色白 呢 ， 胸前 会 有 透 ， 不过 不 影响 穿 着 (This chiffon dress feels heavy when worn. I think summer and spring are suitable for wearing. Putting on this dress will bring out my skin white. But this dress will be transparent on the chest, but it will not affect the wear.)}      \\ \hline
attributes                                    & \multicolumn{1}{p{6.5cm}}{裙长: 中裙||裙型: A 字裙||材质: 涤纶||袖长: 七分 袖||领型: 其它||袖型: 其它||上市时间: 2018 春季||版型: A 型}     \\ \hline
question                                     & \multicolumn{1}{p{6.5cm}}{怀孕 五 个 多 月 能 穿 吗 (I have been pregnant for more than five months, can I wear it?)}   \\ 
\hline
reference                                     & \multicolumn{1}{p{6.5cm}}{能 穿 到 生 都 没 问题 的 (You can wear it until your child is born)}   \\ \hline
S2SA                                          & \multicolumn{1}{p{6.5cm}}{可以 的 ， 我 儿子 三 个 月 就 穿 了 (I can wear it, my son wore it when he was three months old.)}    \\ \hline
RAGF                                          & \multicolumn{1}{p{6.5cm}}{可以 啊 ， 我 的 就 是 五 个 月 (I can wear it, my pregnancy is five months.)}  \\ \hline
PAAG                                          & \multicolumn{1}{p{6.5cm}}{可以 的 ， 我 就是 (I can wear it, I am pregnant.)} \\ 
\bottomrule
\end{tabular}
\vspace{-3mm}
\label{tab:case}
\end{table}
\end{CJK*}

Finally, we address \textbf{RQ4}. Table~\ref{tab:case} shows an example and its corresponding generated answers by different methods.
We observe that S2SA only generates the answer which is fluent, but generated answers are contradictory to the facts.
Due to there is no fact consistency constraint in RAGF, it will also face this problem when generating answers
However, PAAG overcomes this shortcoming by using consistency constraint given by discriminator at training, and then produce the answer which is not only fluent but also consistent with the facts.

\begin{table}[t]
\centering
\caption{Comparison of BLEU scores between different product categories.}
\vspace{-4mm}
\label{tab:comp_domains}
\scriptsize
\begin{tabular}{lllllll}
\toprule
                                           & \multicolumn{2}{c}{PAAG}                                                            & \multicolumn{2}{c}{S2SA}                                                            & \multicolumn{2}{c}{S2SAR}                                                         \\ \cline{2-7} 
                                            & \multicolumn{1}{c}{BLEU1} & \multicolumn{1}{c}{BLEU2}& \multicolumn{1}{c}{BLEU1} & \multicolumn{1}{c}{BLEU2} & \multicolumn{1}{c}{BLEU1} & \multicolumn{1}{c}{BLEU2} \\ \hline
\multicolumn{1}{l|}{Jewelry}                & \textbf{19.53}                   & \multicolumn{1}{l|}{\textbf{6.35}}                     & 17.65                   & \multicolumn{1}{l|}{4.26}                      & 18.74                   & 4.69                 \\
\multicolumn{1}{l|}{Mattress}           & \textbf{18.89}                   & \multicolumn{1}{l|}{4.14}   & 16.35                   & \multicolumn{1}{l|}{3.00}    & 17.52                   & \textbf{5.57}       \\
\multicolumn{1}{l|}{Clothing}               & 18.18                   & \multicolumn{1}{l|}{\textbf{5.17}} & \textbf{18.39}                   & \multicolumn{1}{l|}{4.98}  & 18.36                    & 4.68   \\
\multicolumn{1}{l|}{Kitchenware}            & \textbf{18.00}                   & \multicolumn{1}{l|}{\textbf{4.31}} & 15.23                    & \multicolumn{1}{l|}{3.19} & 17.15                   & 4.09   \\
\multicolumn{1}{l|}{Power and Handtools}     & \textbf{16.34 }                  & \multicolumn{1}{l|}{\textbf{3.98}} & 13.73                   & \multicolumn{1}{l|}{3.20} & 15.60                   & 3.22 \\
\multicolumn{1}{l|}{Skin Care}          & 18.01                   & \multicolumn{1}{l|}{\textbf{4.57}} & 15.39                   & \multicolumn{1}{l|}{3.55}   & \textbf{18.33}                   & 4.40   \\
\multicolumn{1}{l|}{Gardening} & 13.67                   & \multicolumn{1}{l|}{\textbf{2.30}}  & 11.86                   & \multicolumn{1}{l|}{1.52}  & \textbf{15.74}                   & \textbf{2.30}   \\
\multicolumn{1}{l|}{Baby}                   & \textbf{18.22}                   & \multicolumn{1}{l|}{\textbf{4.51}}  & 16.95                   & \multicolumn{1}{l|}{3.71}  & 17.27                   & 3.75 \\
\multicolumn{1}{l|}{Automotive Accessories}       & 17.46                   & \multicolumn{1}{l|}{\textbf{3.43}} & 15.49                    & \multicolumn{1}{l|}{3.14}  & \textbf{17.86}                   & 3.00  \\
\multicolumn{1}{l|}{Gift}                   & \textbf{19.25 }                  & \multicolumn{1}{l|}{3.93}  & 17.23                     & \multicolumn{1}{l|}{3.06}  & 18.39                   & \textbf{4.24} \\
\bottomrule
\end{tabular}
\end{table}

We evaluate performances of PAAG on different categories.
Shown in Table~\ref{tab:comp_domains}, we see that our proposed model beats the other two baselines (S2SA and S2SAR), on majority of product categories in terms of BLEU score.
To prove the significance of the above results, we also do the paired student t-test between our model and baseline methods, the p-value of S2SA is 0.0086 and S2SAR is 0.0100.
From the t-test, we can see that the performance of our model is significantly higher than other baselines.

To investigate the robustness of parameter, we train our model in different parameter size and evaluate them by embedding metric shown in Figure~\ref{fig:parameter_test}.
As the training progresses, the performance of each model is rising.
However, the model with a large number of parameters does not have a great advantage in the final performance of the model with a smaller parameters.

\section{Conclusion}

In this paper, we have proposed the task of product-aware answer generation, which aims to generate an answer for a product-aware question from product reviews and attributes.
To address this task, we have proposed product-aware answer generator (PAAG):
An attention-based question aware review reader is used to extract semantic units from reviews, and key-value memory network based attribute encoder is employed to fuse relevant attributes.
In order to encourage the model to produce answers that match facts, we have employed an adversarial learning mechanism to give additional training signals for the answer generation.
To tackle the shortcomings of vanilla GAN, we have applied the Wasserstein distance as value function in the training of consistency discriminator.
In our experiments, we have demonstrated the effectiveness of PAAG and have found significant improvements over state-of-the-art baselines in terms of metric-based evaluations and human evaluations. 
Moreover, we have verified the effectiveness of each module in PAAG for improving product-aware answer generation.

Future work involves extending our model to multiple hop of memory network used as attribute encoder.

\begin{acks}
We would like to thank the anonymous reviewers for their constructive comments. 
This work was supported by the National Key Research and Development Program of China (No. 2017YFC0804001), the National Science Foundation of China (NSFC No. 61876196, No. 61672058), Alibaba Innovative Research (AIR) Fund. 
Rui Yan was sponsored by CCF-Tencent Open Research Fund and Microsoft Research Asia (MSRA) Collaborative Research Program.
\end{acks}
  
\bibliographystyle{ACM-Reference-Format}
\bibliography{sample-bibliography}

\end{document}